\RequirePackage{fix-cm}

\documentclass{svjour3}                     
\smartqed  

\usepackage{graphicx}
\usepackage{url}
\usepackage{epsfig}
\usepackage{amssymb,amsmath}
\usepackage{color}
\usepackage{textcomp}
\usepackage{microtype}
\usepackage{float}
\usepackage{epstopdf}
\usepackage{longtable}
\usepackage{rotating}
\usepackage{array}
\usepackage{multicol}
\usepackage{multirow}

\begin{document}

\title{Multi-View Product Image Search Using Deep ConvNets Representations}
\author{Muhammet Ba\c{s}tan \and \"{O}zg\"{u}r Y{\i}lmaz}
\authorrunning{Ba\c{s}tan \and Y{\i}lmaz}
\institute{\email{mubastan@gmail.com, yilmazozgur.kaan@gmail.com}}

\maketitle

\begin{abstract}

Multi-view product image queries can improve retrieval performance over single view queries significantly. In this paper, we investigated the performance of deep convolutional neural networks (ConvNets) on multi-view product image search. First, we trained a VGG-like network to learn deep ConvNets representations of product images. Then, we computed the deep ConvNets representations of database and query images and performed single view queries, and multi-view queries using several early and late fusion approaches.

We performed extensive experiments on the publicly available Multi-View Object Image Dataset (MVOD 5K) with both clean background queries from the Internet and cluttered background queries from a mobile phone. We compared the performance of ConvNets to the classical bag-of-visual-words (BoWs). We concluded that (1) multi-view queries with deep ConvNets representations perform significantly better than single view queries, (2) ConvNets perform much better than BoWs and have room for further improvement, (3) pre-training of ConvNets on a different image dataset with background clutter is needed to obtain good performance on cluttered product image queries obtained with a mobile phone.

\end{abstract}

\section{Introduction}
\label{intro}

\begin{figure}[h!]
	\centering
	\includegraphics[width=0.5\textwidth]{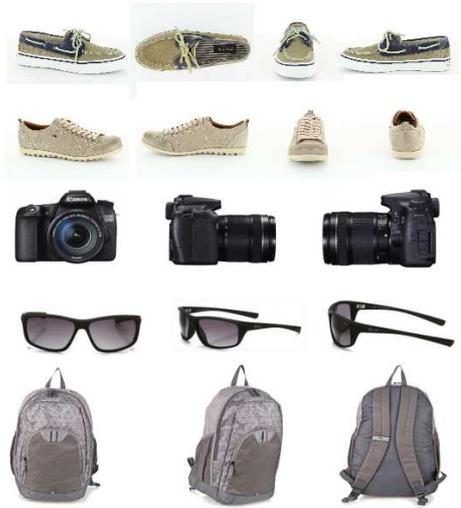}
	\caption{Multi-view images of some products from online shopping sites.}
	\label{fig:multi-view-images}
\end{figure}

Traditionally image retrieval systems are based on single view images of objects in the database and in the query. However, there are some applications in which multi-view images of objects are available in the database. Product image search for online shopping is one such application. Figure~\ref{fig:multi-view-images} shows examples of typical multi-view images of some products from online shopping sites. An image search engine should leverage the availability of such multi-view object image databases to provide more accurate search results to the users.

With the ubiquity of smart phones with cameras, mobile product search is an emerging application area to provide users with an easier and richer shopping experience~\cite{7,14,15}. Users can  easily take one or more photos of a product and search online shopping sites for visually similar products. This motivated the industry to develop mobile visual search applications, such as Google Goggles~\cite{16}, CamFind~\cite{4}, Amazon Flow~\cite{AmazonFlow} and Nokia Point \& Find~\cite{31}.

A mobile product image search system should consider the limitations of the mobile device and use the resources (CPU, memory, battery, network bandwidth) sparingly, while leveraging the user interaction potential to return more accurate results. It is crucial to rank the relevant results within top 10--20 results, because a mobile user will not have time to check a long result list.
Multi-view queries can be helpful in such a context; as the user takes multiple photos of a product, the search system can process each image in the background and communicate with the server for further query processing on the database and return progressively better results using more query images.

Systems using ``multi-image'' queries try to improve retrieval accuracy using multiple images from the same category as the query image~\cite{1,24,49}, the database contains single view images and multiple query images are not multi-view images of the query object. The first work to construct and use a multi-view object image database and multi-view queries is by \c{C}al{\i}\c{s}{\i}r et al.~\cite{mvod-mtap16}. They collected a multi-view product image dataset from online shopping sites, and showed through extensive experiments that multi-view queries on a multi-view database improves retrieval precision significantly compared to single view and multi-image queries. They used bag-of-visual-words (BoWs) as image representation and evaluated many early and late fusion approaches. The presented multi-view query model is independent of the image representations; it is not limited to BoWs. Using deep convolutional neural networks (ConvNets) representations as an alternative to BoWs was set as a future work. Although it has significant potential for product image search, it has not been explored yet.

Deep ConvNets have proven to give state-of-the-art results in many computer vision problems, including image classification, retrieval~\cite{dcnn-iccv2015,deep-retrieval-arxiv16,lcnn-icc2015,local-conv-iccv15,deep-rank-cvpr14} and multi-view recognition~\cite{multi-view-cnn-iccv15}.
In this paper, we investigate the performance of deep ConvNets representations on multi-view product image search, using a multi-view query framework similar to \cite{mvod-mtap16} and compare the performance of the two representations (ConvNets and BoWs).
We also discuss the promising future directions to improve the performance further.

\section{Multi-View Search with Deep ConvNets Representations}
\label{sec:multivs}

In this section, we describe the multi-view search using ConvNets features.
Basically, we use a multi-view query framework similar to~\cite{mvod-mtap16}. We use deep ConvNets representations of product images instead of local feature based BoWs. We also adapt the fusion methods accordingly, using pairwise distances instead of similarities.

A deep convolutional neural network is first trained on labeled image datasets; the labels are object categories.
Then, this ConvNet is run on the database and query images to extract features.
Finally, these ConvNet features are used with Euclidean distance to rank the database objects with early or late fusion, as described below.

\subsection{ConvNet Architecture}
\label{sec:convnet}

\begin{figure}[h!]
	\centering
	\includegraphics[width=\textwidth]{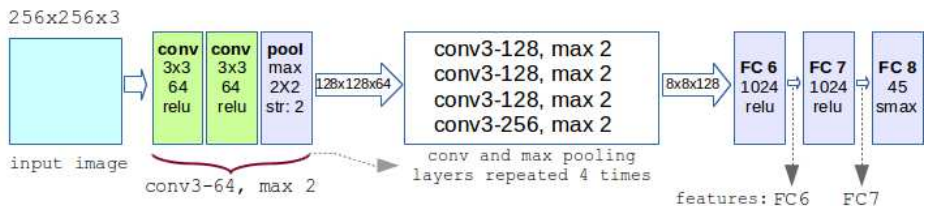}
	\caption{Simplified VGG network architecture used in this work for global image feature extraction.}
	\label{fig:vggnet}
\end{figure}

We use a simplified version of VGG networks~\cite{vgg-iclr15}. The network architecture is shown in Figure~\ref{fig:vggnet}. The input image size is $256\times256\times3$. The network has two convolutional layers with a kernel size of $3\times3$, followed by a max-pooling layer with kernel size $2\times2$ and stride 2; this is repeated five times. At the end of the convolutional layers, the image size is reduced to $8\times8\times128$. The convolutional layers are followed by three fully connected (FC) layers with 1024, 1024 and 45 neurons. Hence, there are 10 $3\times3$ convolutional, 5 $2\times2$ max pooling and 3 fully connected layers in the network.

\begin{figure}[h!]
	\centering
	\includegraphics[width=0.9\textwidth]{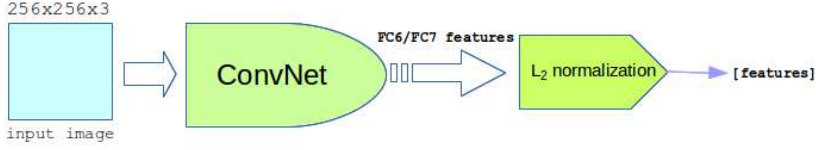}
	\caption{ConvNets feature extraction and $L_2$ normalization.}
	\label{fig:cnnfex}
\end{figure}

The features are extracted as the outputs of the fully connected layers FC6 and FC7. Following common practice, the extracted features are $L_2$ normalized to unit length (Figure~\ref{fig:cnnfex})~\cite{transfer-cvpr15}; this improves the performance significantly.

\subsection{Early Fusion}
\label{sec:earlyfusion}

\begin{figure}[h!]
	\centering
	\includegraphics[width=0.8\textwidth]{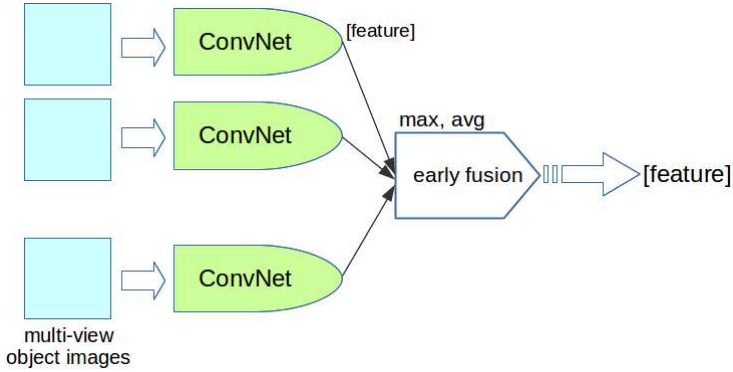}
	\caption{Early fusion.}
	\label{fig:earlyfusion}
\end{figure}

In early fusion, features extracted from multi-view images of an object are reduced to a single feature before applying the distance function (Figure~\ref{fig:earlyfusion}). We used the maximum and average functions (similar to max-pooling and average-pooling) for early fusion. If there are $M$ multi-view images of an object, then the combined feature of an object is computed as

\begin{itemize}
 \item \textit{Maximum (EF-MAX):} $f_i = max(f^1_i, \ldots, f^{M}_i)$
 \item \textit{Average (EF-AVG):} $f_i = \dfrac{\sum_{j=1}^M f^j_i}{M}$
\end{itemize}

where $f_i$ is the $i^{th}$ feature in feature vector $f$. Finally, the distance between the combined feature vectors are measured with Euclidean distance~\cite{transfer-cvpr15} to rank the database objects.

\subsection{Late Fusion}
\label{sec:latefusion}

\begin{figure}[h!]
	\centering
	\includegraphics[width=0.8\textwidth]{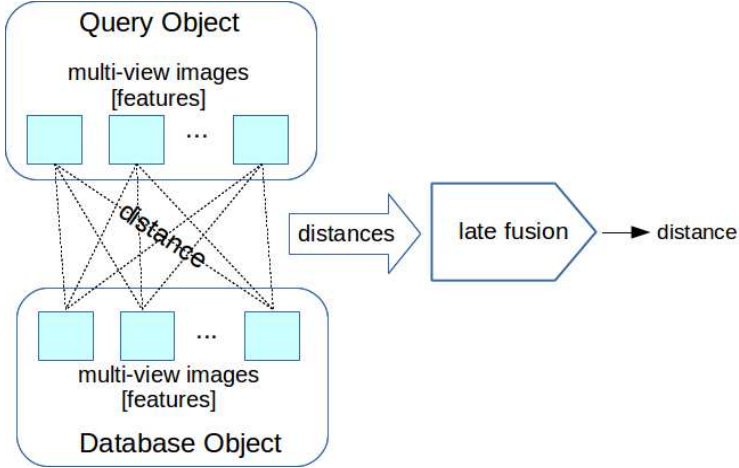}
	\caption{Late fusion.}
	\label{fig:latefusion}
\end{figure}

In late fusion; first, the distances between all pairs of query and database images of an object are computed, then, the distances are combined into a single distance to rank the database objects (Figure~\ref{fig:latefusion}). We use Euclidean distance to measure the dissimilarity between two image features. Once the distances are computed, there are several ways of late fusion: minimum, average, weighted average, etc.

The combined distance measure $d$ between a query object, having $M$ views, and database object, having $N$ views, is computed in one of the following ways in our experiments, based on the results of~\cite{mvod-mtap16} and called \textit{image set distance/similarity}.

\begin{itemize}

 \item \textit{Minimum Distance (LF-MIN):} The distance between a query and database object is taken as the minimum of all $M \times N$ distances.
    
    \begin{equation*}
      \displaystyle
	d = min(d_{ij})
    \end{equation*}
 
 \item \textit{Average Distance (LF-AVG):} The distance is computed as the average of all $M \times N$ distances.
      \begin{equation*}
	\displaystyle
	  d = \dfrac{{\displaystyle\sum^M_{i=1}}~{\displaystyle\sum^N_{j=1}}~{d_{ij}}}{M \times N}
      \end{equation*}

 \item \textit{Weighted Average Distance (LF-WAVG, LF-IWAVG):} A weight $w_{ij}$ is assigned to each distance $d_{ij}$. This weight can be proportional or inversely proportional ($1/d_{ij}$) to the distance.

	    \begin{equation*}
		\displaystyle
		\begin{split}		
		    w_{ij} &= \dfrac{d_{ij}}{ {\displaystyle\sum^M_{i=1}}~{\displaystyle\sum^N_{j=1}}~{d_{ij}} }\\		
		    d &= {\sum^M_i}~{\sum^N_j}~{d_{ij}}\times{w_{ij}} 
		\end{split}
	    \end{equation*}
 
 \item \textit{Average of Minimum Distances (LF-MIN-AVG):} First, the minimum distance for each of $M$ query images to $N$ database object images is computed. Then, the average of $M$ minimum distances is computed as the image set distance.
 
	    \begin{equation*}
	      \displaystyle
	      d = \dfrac{ \sum^M_i {\min (d_{i1}, \ldots,d_{iN} ) } } {M}
	    \end{equation*}

 \item \textit{Weighted Average of Minimum Distances (LF-MIN-WAVG):} This is the weighted average version of the previous method; the weighting is proportional.
 
      \begin{equation*}
	    \displaystyle
	    \begin{split}
	    d_i    &= \max (d_{i1}, \ldots,d_{iN} )\\
	    w_{i} &= \dfrac{d_{i}}{ {\sum^M_i}{d_{i}} }\\
	    \displaystyle
	    d &= {\sum^M_i}{w_i}\times{d_i} 
	    \end{split}
      \end{equation*}

\end{itemize}

\section{Dataset and Evaluation}
\label{sec:data}

\begin{figure}[h!]\centering
 	\fbox{\includegraphics[width=0.60\textwidth]{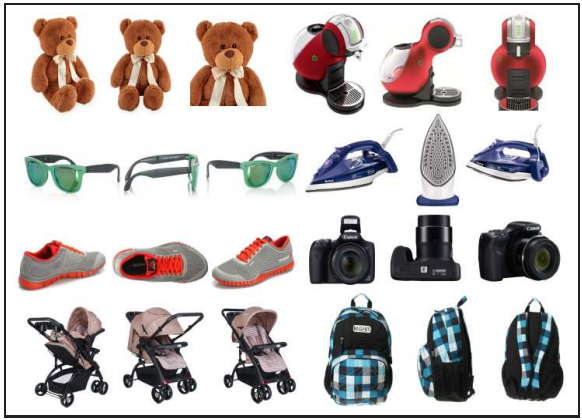}}
	\caption{Sample images from the MVOD 5K database.}
	\label{fig:mvod-dataset}
\end{figure}

We used the Multi-View Object Image Dataset (MVOD 5K)~\cite{mvod-mtap16}, publicly available at \url{www.cs.bilkent.edu.tr/~bilmdg/mvod/}.
The database has 5K images from $45$ different product categories (shoes, backpacks, eyeglasses, cameras, printers, guitars, coffee machines, etc.).
Each object has at least two different images taken from different views (multi-view).
The images mostly have a clean background and objects are positioned at the image center. Figure~\ref{fig:mvod-dataset} shows sample images from the database.

\begin{figure}[h!]\centering
 	\fbox{\includegraphics[width=0.60\textwidth]{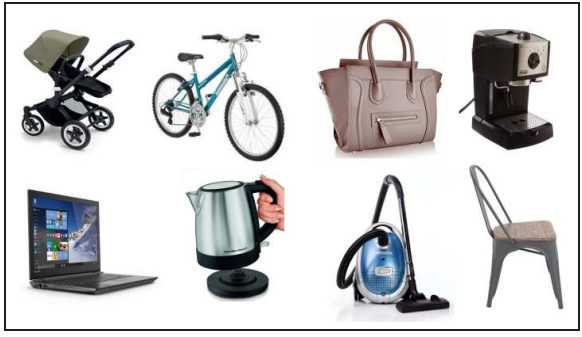}}
	\caption{Sample images from the MVOD Internet queries. All queries have multi-view images; here, only a single view image is shown for the sample queries.}
	\label{fig:internet-query}
\end{figure}

\begin{figure}[h!]\centering
 	\fbox{\includegraphics[width=0.60\textwidth]{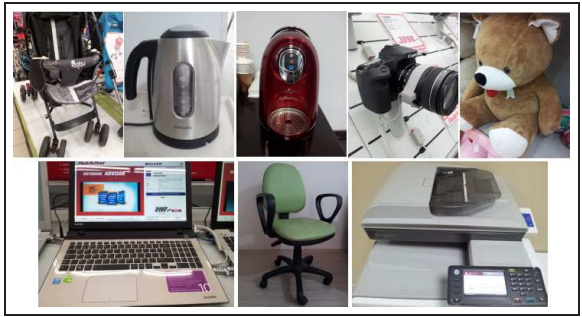}}
	\caption{Sample images from the MVOD phone queries. All queries have multi-view images; here, only a single view image is shown for the sample queries.}
	\label{fig:phone-query}
\end{figure}

In addition to the 5K multi-view database images, the MVOD dataset has two sets of multi-view object queries:

\begin{description}

 \item[\textbf{Internet Queries.}] The multi-view query images are collected from online shopping sites, similar to the MVOD 5K database (Figure~\ref{fig:internet-query}). There are $45$ queries in the set, one query per object category. The images are similar to MVOD 5K, they mostly have clean background and objects are positioned at the image center.
 
 \item[\textbf{Phone Queries.}] The multi-view query images are collected with a mobile phone in natural office, home or supermarket environments (Figure~\ref{fig:phone-query}). The query set has $15$ queries for $15$ categories. The phone queries are more difficult than the Internet queries, since they have adverse effects, like background clutter and illumination problems.

\end{description}

We evaluated the retrieval performance in terms of average precision (AveP), as in~\cite{mvod-mtap16,34}. The average precision is calculated as shown below; \textit{k} represents the rank in the result list and \textit{N} is the length of the result list.

  \begin{equation*}
  \begin{split}
  \label{eq:AvgPrec}
    P(k) = \dfrac{\mbox{relevant objects} {\;\cap\;} \mbox{first k objects}}{k}\\
    rel(k)= \begin{cases}
    1, &\mbox{if object k is relevant} \\
    0, & \mbox{otherwise}
    \end{cases} \\
    \displaystyle
    AveP = \dfrac{{\sum_{k=1}^{N}}({P(k){\times}rel(k)})}{N}
  \end{split}
  \end{equation*}

Queries and query results are object-based; a list of objects are returned as the query result. A retrieved object is relevant (correct) if it belongs to the same object category as the query object~\cite{mvod-mtap16,34}. Although the evaluation is not based on instance retrieval, i.e., returning objects that are instances of the query object, it is desirable to rank visually similar database objects higher in the result list. There is no such publicly available multi-view object image dataset to use for instance-based retrieval evaluation.

The interpolated average precision graphs are obtained by averaging the average precisions over all queries.

\section{Experiments}
\label{sec:experiments}

We performed extensive single and multi-view retrieval experiments on the MVOD 5K dataset and evaluated the performance of various fusion methods for multi-view queries. We also compared performance of ConvNets features to BoWs~\cite{mvod-mtap16}, which is the only existing work on multi-view product image search.

We implemented the ConvNets training and feature extraction using Google's TensorFlow library~\cite{tensorflow} and a higher level API TFLearn~\cite{tflearn}. The categories in the dataset are mutually exclusive, only one object category is labeled in each image. We used one-hot encoding for the labels and minimized categorical cross entropy loss with Adam optimizer, which is available in TensorFlow. The ConvNet was first trained on Caltech 256 Object Categories dataset~\cite{caltech256}, with 20K images for 200 epochs, then refined on MVOD 5K database for 50 epochs. The images are resized to $256\times256\times3$, with padding for non-square images. Data augmentation (Gaussian blur with a maximum $\sigma$ of 2.0, random left-right flip) is also employed.

After training, features are extracted from the database and query images using the fully connected layer outputs (FC6 and FC7) and $L_2$ normalized. Finally, the extracted global image features are used for retrieval with Euclidean distance. FC6 and FC7 features result in similar performance; the results below were obtained with FC7 features since they were slightly better. Feature extraction takes about 0.12 seconds per $256\times256\times3$ image, with TensorFlow on NVIDIA GEFORCE GT 650M GPU with 2GB memory, Intel CORE i7 2.4GHz CPU.

\subsection{Results on the Internet Queries}
\label{sec:internet-result}

\begin{figure}[h!]\centering
 	\fbox{\includegraphics[width=0.70\textwidth]{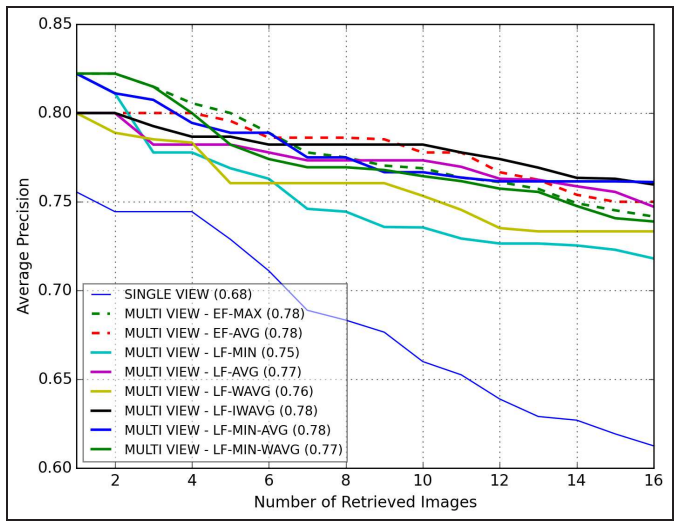}}
	\caption{Interpolated average precision graphs for 45 Internet queries on MVOD 5K. Numbers inside the parenthesis are the mean average precisions (mAP).}
	\label{fig:avep-internet-query}
\end{figure}

Figure~\ref{fig:avep-internet-query} shows interpolated average precision graphs for 45 Internet queries on the MVOD 5K database. Similar to the results of~\cite{mvod-mtap16}, multi-view queries with early and late fusion are significantly better than single view queries. Late fusion methods give similar results, with averaging methods being slightly better. Based on these results, early fusion approaches (maximum, average) can be preferred, since early fusion methods are faster than late fusion methods (with unoptimized, exhaustive search, single threaded Python implementation, early fusion maximum takes 0.132 seconds, early fusion average takes 0.148 seconds and late fusion average takes 0.319 seconds per query, on an Intel CORE i7 2.4GHz laptop with 32GB memory). Concatenating the FC6 and FC7 features did not result in meaningful improvement.

A network trained only on Caltech 256~\cite{caltech256} dataset for 200 epochs, without any refinement on MVOD 5K, performed rather poorly on the Internet queries (about half the performance). On the other hand, network trained only on MVOD 5K dataset performed similarly to the presented results. This may be because the Internet queries are very similar to the database.

\subsection{Results on the Phone Queries}
\label{sec:phone-result}

\begin{figure}[h!]\centering
 	\fbox{\includegraphics[width=0.70\textwidth]{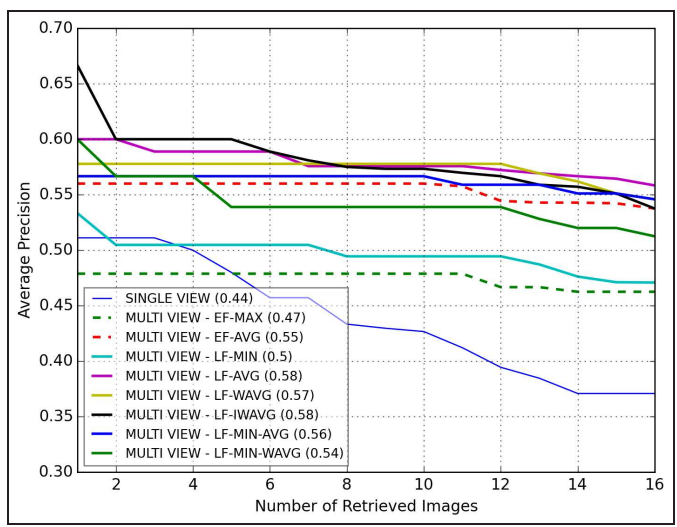}}
	\caption{Interpolated average precision graphs for 15 phone queries on MVOD 5K. Numbers inside the parenthesis are the mean average precisions (mAP).}
	\label{fig:avep-phone-query}
\end{figure}

Figure~\ref{fig:avep-phone-query} shows interpolated average precision graphs for 15 phone queries on the MVOD 5K database. As before, multi-view queries are significantly better than single view queries. Late fusion methods are better than early fusion, and averaging late fusion methods are slightly better than other late fusion methods. Concatenating the FC6 and FC7 features did not result in meaningful improvement.

A network trained only on Caltech 256~\cite{caltech256} dataset for 200 epochs, without any refinement on MVOD 5K, performed rather poorly on the phone queries as well (about half the performance). Moreover, a network trained only on MVOD 5K also performed poorly. This can be explained by the fact that, the database images have mostly clean background without any illumination problems, while the phone queries have cluttered backgrounds with adverse illumination effects. The network should be trained to account for such adverse effects.

\subsection{ConvNets versus BoWs}
\label{sec:cnn-bow}

In this section, we compare the retrieval performance of ConvNets features to the BoWs of~\cite{mvod-mtap16}, on both the Internet and phone queries. In~\cite{mvod-mtap16}, Harris and Hessian keypoint detectors with SIFT descriptors were used with a vocabulary size of $3K$ and hard assignment. The BoWs of Harris and Hessian were concatenated and a BoW histogram of size $6K$ was obtained; this is usually a very sparse histogram. Various similarity functions were evaluated to measure the similarity between BoW histograms and \textit{Min-Max Ratio} was found to be the best. The BoW results given here (the best results reported in~\cite{mvod-mtap16}) use the \textit{Min-Max Ratio} function.

\begin{figure}[h!]\centering
 	\fbox{\includegraphics[width=0.70\textwidth]{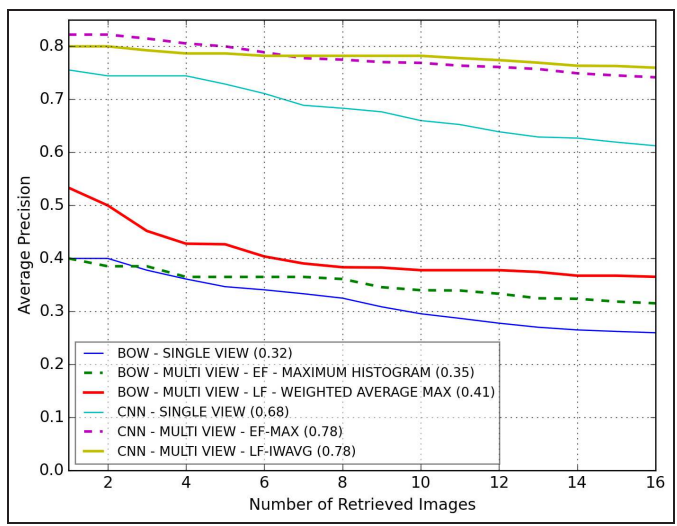}}
	\caption{Comparison of CNN and BoW features on 45 Internet queries, on MVOD 5K database. Numbers inside the parenthesis are the mean average precisions (mAP).}
	\label{fig:avep-cnn-bow-internet-query}
\end{figure}

\begin{figure}[h!]\centering
 	\fbox{\includegraphics[width=0.70\textwidth]{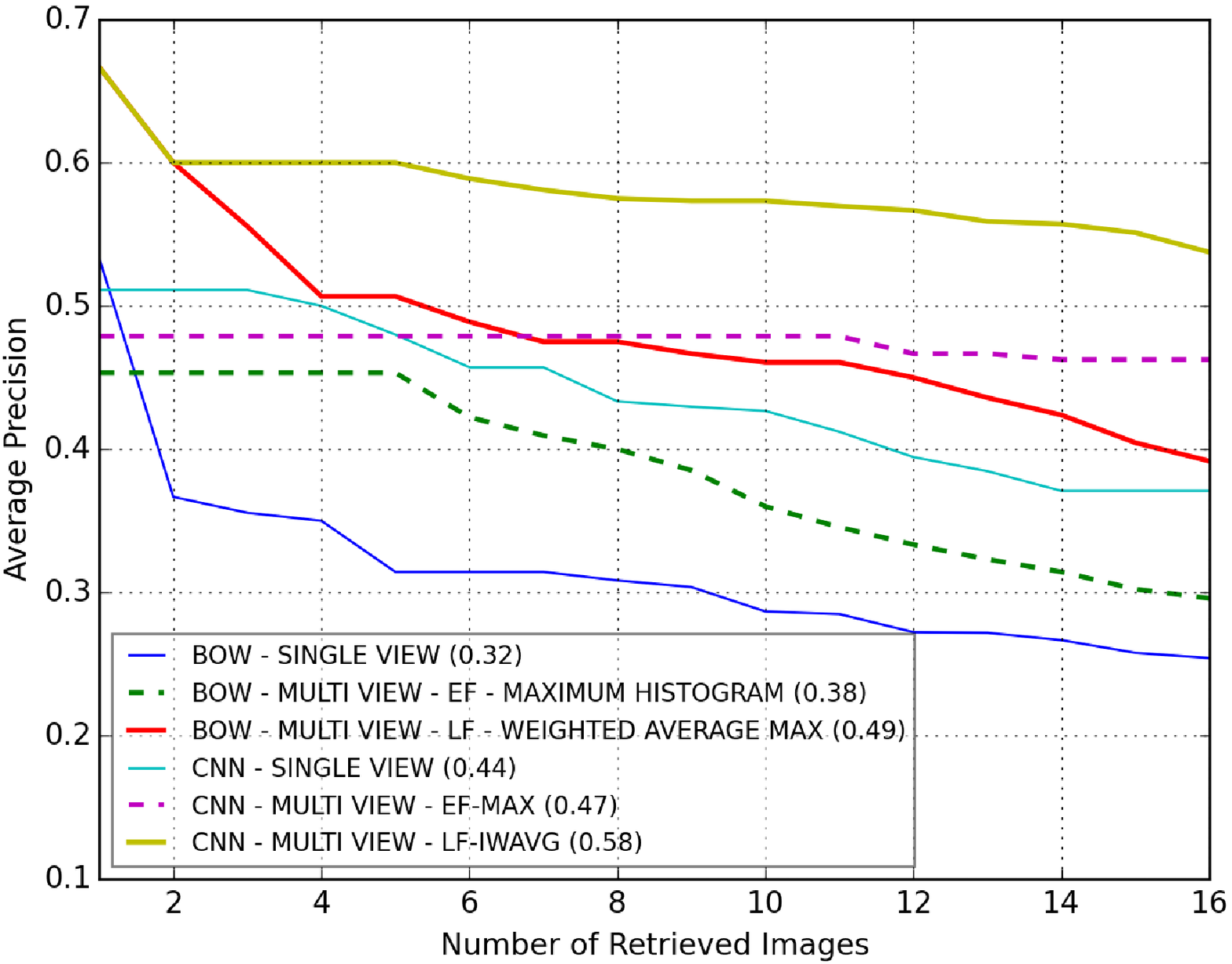}}
	\caption{Comparison of CNN and BoW features on 15 phone queries, on MVOD 5K database. Numbers inside the parenthesis are the mean average precisions (mAP).}
	\label{fig:avep-cnn-bow-phone-query}
\end{figure}

\begin{figure}[h!]\centering
 	\fbox{\includegraphics[width=0.70\textwidth]{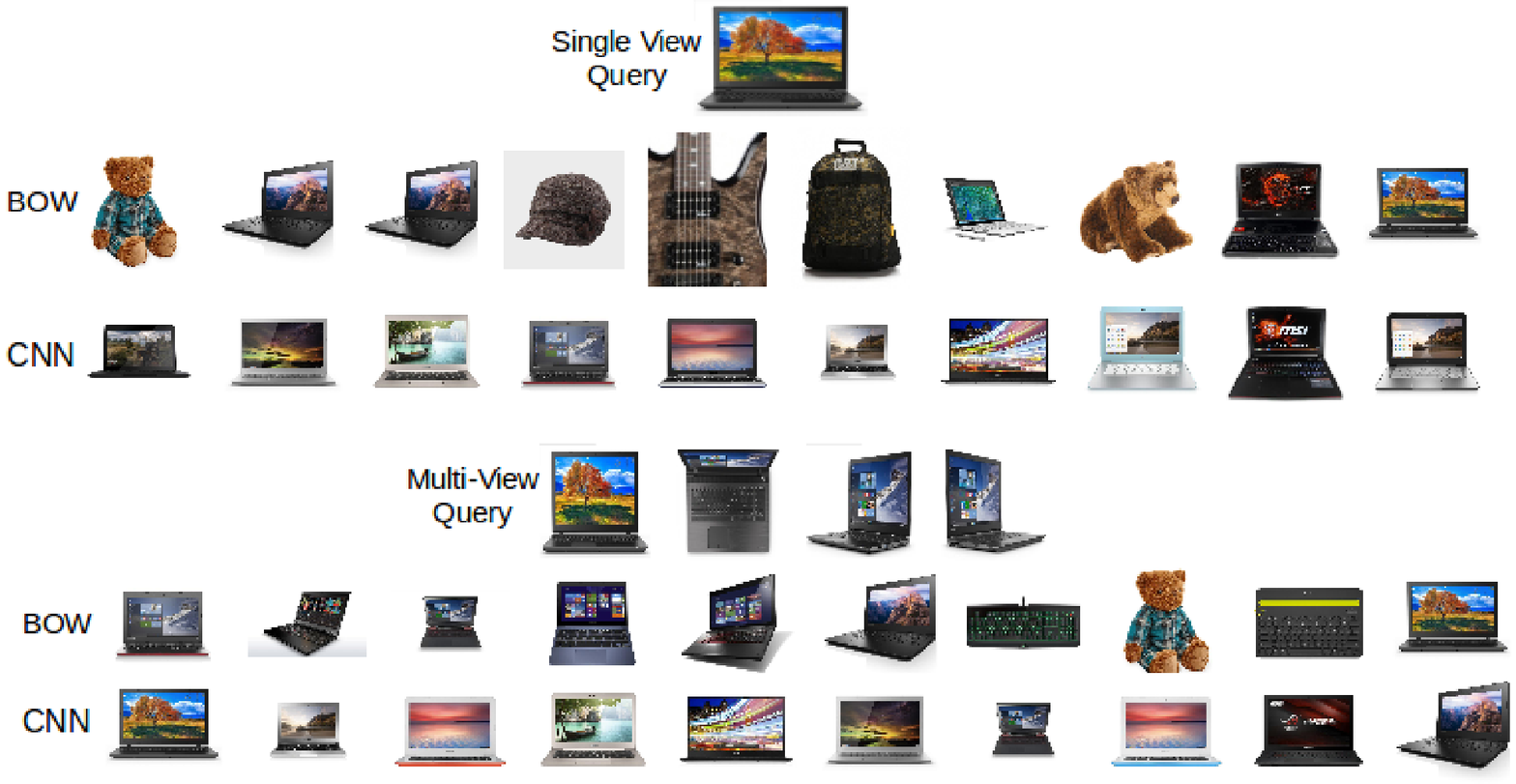}}
	\caption{Sample single and multi-view Internet query results with BoWs and ConvNets.}
	\label{fig:sample-query-internet1}
\end{figure}

\begin{figure}[h!]\centering
 	\fbox{\includegraphics[width=0.70\textwidth]{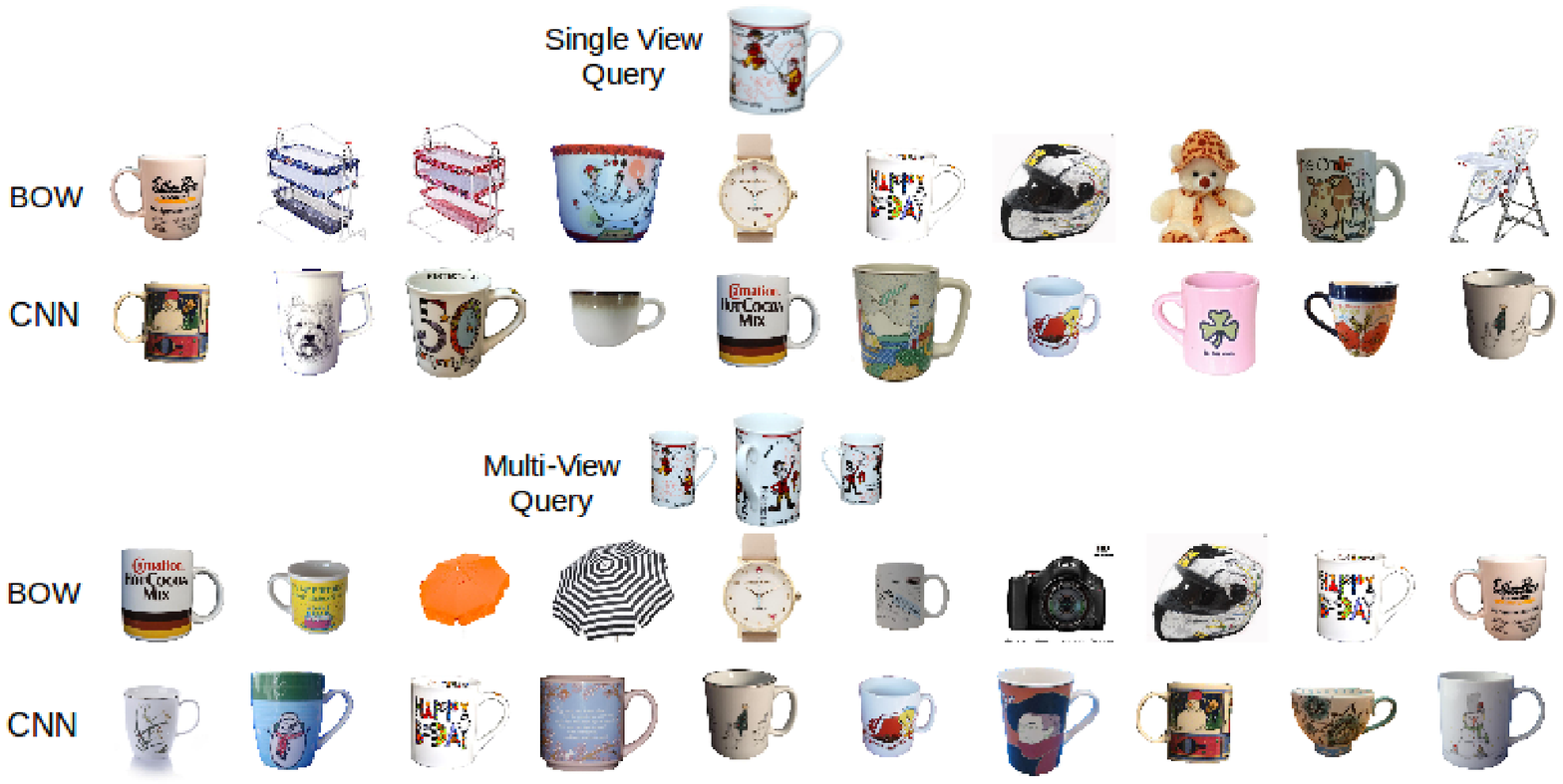}}
	\caption{Sample single and multi-view Internet query results with BoWs and ConvNets.}
	\label{fig:sample-query-internet2}
\end{figure}

\begin{figure}[h!]\centering
 	\fbox{\includegraphics[width=0.70\textwidth]{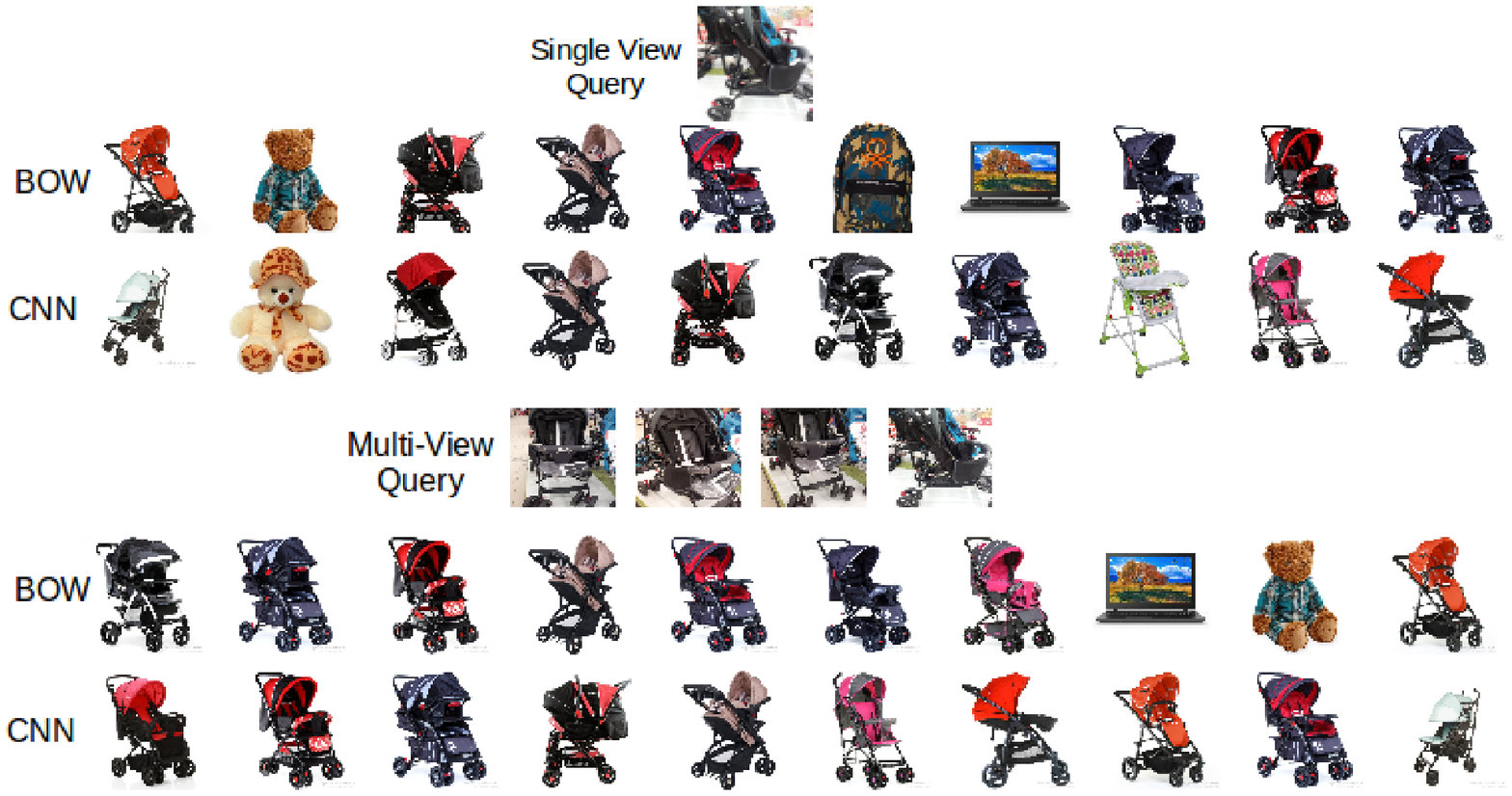}}
	\caption{Sample single and multi-view phone query results with BoWs and ConvNets.}
	\label{fig:sample-query-phone1}
\end{figure}

\begin{figure}[h!]\centering
 	\fbox{\includegraphics[width=0.70\textwidth]{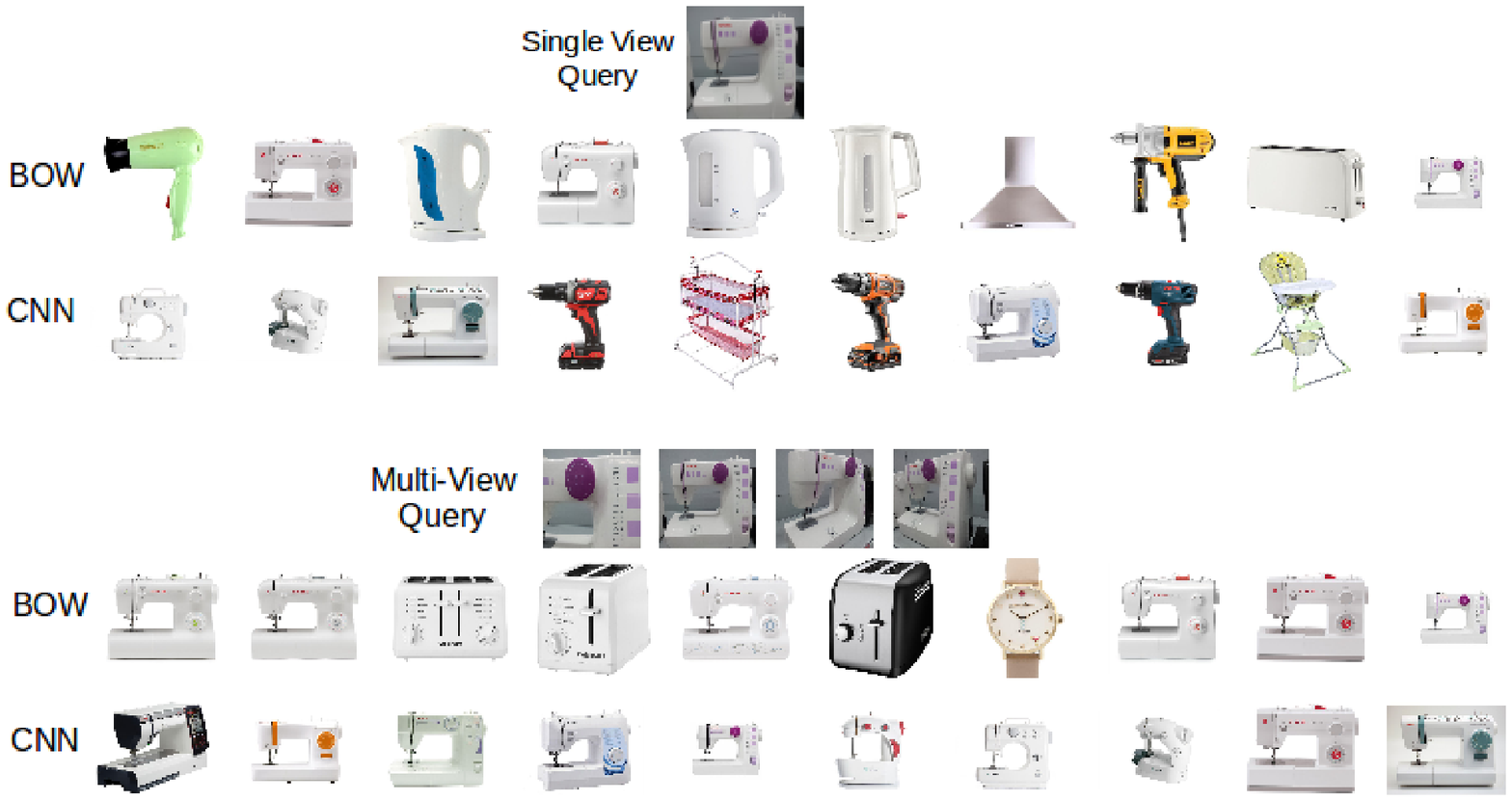}}
	\caption{Sample single and multi-view phone query results with BoWs and ConvNets.}
	\label{fig:sample-query-phone2}
\end{figure}

Figures~\ref{fig:avep-cnn-bow-internet-query} and~\ref{fig:avep-cnn-bow-phone-query} show the interpolated average precision graphs for ConvNets and BoWs on the Internet and phone queries, respectively. Both single view and multi-view average precision values of ConvNets are significantly higher on both query types. The improvement is especially huge on the Internet queries, almost doubling the average precision. Moreover, the dimensionality of ConvNets features (1024) is much lower than that of BoWs (6K). Figures~\ref{fig:sample-query-internet1}--\ref{fig:sample-query-phone2} show sample Internet and phone query results with BoWs and ConvNets. The samples are selected to be the same as those of~\cite{mvod-mtap16} for comparison. 

\section{Conclusions and Future Directions}
\label{sec:conclusion}

We investigated the performance of deep ConvNets representations on multi-view product image search using various early and  late fusion methods. Similar to the results of~\cite{mvod-mtap16}, multi-view queries on multi-view database result in significant performance improvement in terms of average precision.
We also compared the ConvNets features with classical BoWs~\cite{mvod-mtap16} and found ConvNets to be much better.
Moreover, there is still room for performance improvement with ConvNets, by designing better networks and training on larger datasets.

This work did not analyze the applicability of the designed network to mobile product search systems, which have memory and processing power limitations.
It is left as a future work to design high performance ConvNets suitable for mobile product search systems; SqueezeNet-like networks~\cite{SqueezeNet-arxiv16,deep-quantization-aaai16} with compression is a promising direction.
Another direction is to build larger datasets for better training of ConvNets and more realistic performance evaluation.

Finally, with a large enough dataset, it may be possible to improve performance further using the end-to-end training of three-stream siamese networks with triplet loss for instance-level retrieval~\cite{rmac-triplet-2016} and also training directly on multiple views instead of training on single views and fusing at retrieval.

\bibliographystyle{spmpsci}      
\bibliography{References}

\end{document}